\documentclass{article}

\usepackage{arxiv}

\usepackage[utf8]{inputenc} % allow utf-8 input
\usepackage[T1]{fontenc}    % use 8-bit T1 fonts
\usepackage{hyperref}       % hyperlinks
\usepackage{url}            % simple URL typesetting
\usepackage{booktabs}       % professional-quality tables
\usepackage{amsfonts}       % blackboard math symbols
\usepackage{nicefrac}       % compact symbols for 1/2, etc.
\usepackage{microtype}      % microtypography

\usepackage{graphicx}
\usepackage{natbib}
\usepackage{doi}

\usepackage{amsthm}
\theoremstyle{definition}
\newtheorem{definition}{Definition}[section]
\usepackage{ulem}
\usepackage{standalone}
\usepackage{pifont}
\usepackage{tikz}
\usepackage{subcaption}
\usepackage{enumitem}
\setlist[enumerate]{label*=\arabic*.}
\usetikzlibrary{shapes}
\usepackage{amsmath}

\title{Why Should I Choose You?\\\large{AutoXAI: A Framework for Selecting and Tuning eXplainable AI Solutions}}

\author{ \href{https://orcid.org/0000-0001-9316-0617}{\includegraphics[scale=0.06]{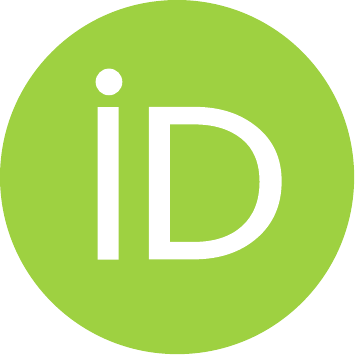}\hspace{1mm}Robin Cugny}
    \\
	IRIT\\
	Université Toulouse 2\\
	Toulouse, France \\
	\texttt{robin.cugny@irit.fr} \\
	\And
	\href{https://orcid.org/0000-0002-1954-8733}{\includegraphics[scale=0.06]{orcid.pdf}\hspace{1mm}Julien Aligon} \\
	IRIT\\
	Université Toulouse 1\\
	Toulouse, France \\
	\texttt{julien.aligon@irit.fr} \\
	\And
	\href{https://orcid.org/0000-0001-5402-6255}{\includegraphics[scale=0.06]{orcid.pdf}\hspace{1mm}Max Chevalier} \\
	IRIT\\
	Université Toulouse 3\\
	Toulouse, France \\
	\texttt{max.chevalier@irit.fr} \\
	\And
	\href{https://orcid.org/0000-0002-2355-5901}{\includegraphics[scale=0.06]{orcid.pdf}\hspace{1mm}Geoffrey Roman Jimenez} \\
	SolutionData Group\\
	Toulouse, France \\
	\texttt{groman-jimenez@solutiondatagroup.fr} \\
	\And
	\href{https://orcid.org/0000-0003-0338-9886}{\includegraphics[scale=0.06]{orcid.pdf}\hspace{1mm}Olivier Teste} \\
	IRIT\\
	Université Toulouse 2\\
	Toulouse, France \\
	\texttt{olivier.teste@irit.fr} \\
}

\renewcommand{\shorttitle}{Why Should I Choose You?}

\hypersetup{
pdftitle={Why Should I Choose You?},
pdfsubject={cs.LG},
pdfauthor={Robin Cugny, Julien Aligon, Max Chevalier, Geoffrey Roman Jimenez, Olivier Teste},
pdfkeywords={Explainable machine learning, Evaluation of explainability, Quality of explanation, Evaluation metrics, AutoML, Recommender system, Information system},
}

\begin{document}
\maketitle

\begin{abstract}
In recent years, a large number of XAI (eXplainable Artificial Intelligence) solutions have been proposed to explain existing ML (Machine Learning) models or to create interpretable ML models. Evaluation measures have recently been proposed and it is now possible to compare these XAI solutions. However, selecting the most relevant XAI solution among all this diversity is still a tedious task, especially when meeting specific needs and constraints.
In this paper, we propose AutoXAI, a framework that recommends the best XAI solution and its hyperparameters according to specific XAI evaluation metrics while considering the user's context (dataset, ML model, XAI needs and constraints). It adapts approaches from context-aware recommender systems and strategies of optimization and evaluation from AutoML (Automated Machine Learning). 
We apply AutoXAI to two use cases, and show that it recommends XAI solutions adapted to the user's needs with the best hyperparameters matching the user's constraints.
\end{abstract}

% keywords can be removed
\keywords{Explainable machine learning, Evaluation of explainability, Quality of explanation, Evaluation metrics, AutoML, Recommender system, Information system}

\section*{Acknowledgements}
This work was supported by ANRT (CIFRE [2020/0870]) in collaboration with SolutionData Group and IRIT.

\section{Introduction}\label{section:Intro}
Machine Learning (ML) models are now widely used in the industry. However, their lack of understandability delays their adoption in high stakes domains such as the medical field \cite{markus2020role}, digital security \cite{brown2018recurrent}, judicial field \cite{tan2018distill}, or autonomous driving \cite{9616449}. 
In such contexts, decision-makers should understand ML models and their results to detect biases \cite{tan2018distill} or meaningless relationships \cite{ribeiro2016should}. During the last decade, the eXplainable Artificial Intelligence (XAI) field proposed a wide variety of solutions to facilitate the understanding of ML models \cite{adadi2018peeking,carvalho2019machine, molnar2020interpretable, arrieta2020explainable,doshi2017towards,lipton2018mythos,gilpin2018explaining}. In view of the growing number of XAI proposals \cite{arrieta2020explainable}, evaluating the quality of explanations has become necessary to choose an appropriate XAI solution as well as its hyperparameters. It is worth noting that the evaluation of explanations can either be done subjectively by humans or objectively with metrics \cite{nguyen2020quantitative,zhou2021evaluating,nauta2022anecdotal}. 
However, data scientists who want to include an XAI solution have the following issues:
\begin{itemize}
    \item They must check which XAI solutions are compatible with the data type and the ML model.
    \item The XAI solutions should explain specifically what the data scientists want to understand and it should be explained in an appropriate format.
    \item They should evaluate the effectiveness of the explanations produced by the selected XAI solutions.
    \item The context requires that the explanations match specific quality criteria (called explanations' properties) which imposes the use of appropriate evaluation metrics.
    \item They have to find the best hyperparameters for each of the selected XAI solutions to keep the best of them.
\end{itemize}
These are tedious and time-consuming tasks. Theoretical guides have been proposed by \cite{Axa,Palacio_2021_ICCV} but, to the best of our knowledge, automating the complete XAI recommendation approach has never been formalized and implemented before.

In this paper, we propose to automate these tasks in a framework to assist data scientists in choosing the best XAI solutions according to their context (dataset, ML model, XAI needs and constraints).

Suggesting an adapted XAI solution requires defining the elements of the data scientists' contexts and using them to filter the compatible XAI solutions. This task is challenging because there are as many formalizations as there are authors in the XAI field and very few works have attempted to unify XAI elements with a formalization \cite{lundberg2017unified,Palacio_2021_ICCV,nauta2022anecdotal}. As we want to evaluate the XAI solutions, we must automatically find the XAI evaluation metrics that are compatible with the XAI solutions and are meaningful to the context. Moreover, it is necessary to find a way to validate multiple complementary properties by optimizing corresponding XAI evaluation metrics while considering the data scientists' preferences. Indeed, properties' importance is subjective and depend on the context. In addition, ranking XAI solutions requires finding the best hyperparameters using XAI evaluation metrics. As it is computationally expensive, we draw inspiration from time-saving strategies for model evaluation in AutoML \cite{he2021automl}.

The contributions of this paper are as follows:
\begin{itemize}
    \item AutoXAI, a framework that recommends XAI solutions to match the data scientists' context and optimize their hyperparameters with respect to XAI evaluation metrics \cite{zhou2021evaluating,nauta2022anecdotal}.
    \item A more generic formalization of the data scientists' context for XAI.
    \item A new evaluation metric to assess the completeness of example-based explanations.
    \item New time-saving strategies adapted to XAI evaluation.
\end{itemize}
We illustrate AutoXAI's recommendations through two use cases with different users' constraints and needs as well as different datasets and models. These studies let us uncover interactions between hyperparameters and properties of explanations, as well as interactions between the properties themselves.

The rest of the paper is organized as follows. 
Related work are described in Section \ref{section:Related}. Formal definitions are illustrated by an example of context in Section \ref{section:exandformal}. The core of our framework is detailed in Sections \ref{section:framework} and \ref{section:time}. Experiments of Section \ref{section:experiments} show that our framework is well adapted to propose an explanation matching the user's context and that time-saving strategies considerably reduce the computation time. Finally, we conclude the paper and give possible perspectives in Section \ref{section:Conclusion}.

\section{Related work}\label{section:Related}

Four research topics interact in this paper: the XAI solutions, which are being assessed by XAI evaluation metrics while context-aware recommender systems and AutoML are means used to propose the most adapted solution.

\subsection{XAI solutions}\label{subsection:XAI}
In this paper, we define an XAI solution as any algorithm that produces an explanation related to an ML problem. This includes methods that explain black-box models but also naturally interpretable models. As mentioned in Section \ref{section:Intro}, many XAI solutions now exist and different taxonomies have been proposed such as \cite{arrieta2020explainable, carvalho2019machine,adadi2018peeking,gilpin2018explaining,nauta2022anecdotal}.
\cite{carvalho2019machine} also suggests grouping XAI solutions according to the type of explanation produced. They list: feature summary, model internals, data point, surrogate intrinsically interpretable model, rule sets, explanations in natural language, and question-answering. Later, \cite{liao2020questioning} suggests that XAI explanations answer specific questions about data, its processing and results in ML. They map existing XAI solutions to questions and create an XAI question bank that supports the design of user-centered XAI applications.
\cite{overton2011scientific} defines an explanation as an explanan: the answer to the question and an explanandum: what is to be explained. These two elements provide a user-friendly characterization of explanations and thus allow the user to specify which explanation is more adapted.

The diversity of existing XAI solutions makes it hard to find an XAI solution adapted to one's needs. Moreover, as the XAI field is growing, more and more XAI solutions proposed in the literature are producing similar kinds of explanations. Hence, it has become necessary to objectively compare XAI solutions by assessing the effectiveness of their explanations. In this direction, the recent literature has focused on quantitative XAI evaluations \cite{nauta2022anecdotal}.

\subsection{Evaluation of XAI solutions}\label{subsection:evaluation}

\cite{doshi2017towards} distinguishes three strategies of evaluation: application-grounded evaluation, human-grounded evaluation, and functionality-grounded evaluation that does not imply human intervention.
Application-grounded evaluation tests the effectiveness of explanations in a real-world application with domain experts and human-grounded evaluation are carried out with lay humans.
While explanations are intended for humans, functionality-grounded evaluations are interesting because of their objectivity. Thus, this type of evaluation is inexpensive, fast and can lead to a formal comparison of explanation methods \cite{zhou2021evaluating}. 

Since the notion of "good explanations" is not trivial, some quality properties have been proposed by \cite{robnik2018perturbation}. These are man-made criteria that attest to the quality of the explanations. Functionality-grounded evaluation metrics are constructed to calculate scores to measure how well a property is met.

\cite{nauta2022anecdotal} focuses on the functionality-grounded evaluation and proposed the Co-12 Explanation Quality Properties to unify the diverse properties proposed in the literature. They reviewed most XAI evaluation metrics and associate each of them with properties. Examples of their properties that will be studied in this paper are as follow: \textit{Continuity} describes how continuous and generalizable the explanation function is, \textit{Correctness} describes how faithful the explanation is w.r.t. the black box, \textit{Compactness} describes the size of the explanation, and \textit{Completeness} describes how much of the black box behavior is described in the explanation.

In practice, XAI evaluation metrics produce scores for properties of interest, making it possible to compare and choose an XAI solution. However, the data scientists still have to find the desired XAI solutions and their corresponding XAI evaluation metrics. This issue could be addressed with strategies that have been studied in context-aware recommender systems.

\subsection{Context-aware recommender systems}\label{subsection:CARS}
Recommender systems filter information to present the most relevant elements to a user. To the best of our knowledge, there is no recommender system for XAI solutions. To recommend adapted XAI solutions, one should consider the whole context of the data scientist. According to \cite{adomavicius2011context}, context-aware recommender systems offer more relevant recommendations by adapting them to the user's situation. They also state that context may be integrated during three phases: contextual prefiltering which selects a subset of possible candidates before the recommendation, contextual modeling which uses context in the recommendation process, and contextual postfiltering which adjusts the recommendation afterward. These three phases require formally defining the elements of the context, which is one of our objectives for the framework we propose in this paper. While recommending an adapted XAI solution is a first interesting step, the data scientist eventually wants a reliable explanation, i.e. an explanation that verifies the properties of interest. To achieve this, a possible approach is to use previously detailed XAI evaluation metrics to optimize hyperparameters of adapted XAI solutions. For this kind of approach, many strategies have been proposed in the AutoML domain.

\subsection{AutoML}\label{subsection:automl}
Designing ML algorithms is an iterative task of testing and modifying both the architecture and the hyperparameters of the algorithm. It is a repetitive task that requires a lot of time. For this reason, a part of the research has focused on automating the design of ML algorithms, namely AutoML \cite{he2021automl}. AutoML frameworks look for the best performing ML pipeline treatment to solve a task on a given dataset. According to \cite{he2021automl}, AutoML consists of several processes: data preparation, feature engineering, model generation, and model evaluation. They divide the model generation process into two steps: search space and optimization methods. The first step defines the design principles of models that are tested, and the second is how to obtain the best scores in the model evaluation process. The main strategy of interest here is HyperParameter Optimization (HPO) which consists in finding the best hyperparameters according to a loss function. As model performances cannot be derived according to a hyperparameter, it is a non-differentiable optimization problem, therefore the HPO methods do not rely on the model to propose a solution.  
Moreover, since training a model until convergence is a costly operation, model evaluation is a very time-consuming step. Thus, several strategies have been proposed to accelerate the evaluation of models.
\cite{he2021automl} lists four types of strategies: \textit{low fidelity}, \textit{weight sharing}, \textit{surrogate}, and \textit{early stopping}. The \textit{low fidelity} strategy consists in reducing the number of observations to reduce the number of epochs, or the size of observations to reduce the number of parameters to optimize. \textit{Weight sharing} reuses learned parameters between models. The \textit{surrogate} strategy replaces a computationally expensive model with an approximation to estimate the performance of neural architecture and guide architecture research. \textit{Early stopping} can accelerate model evaluation by stopping iterations if performances are predicted to be lower than the current best score.

\subsection{Approaches for choosing an XAI solution}
As mentioned in Section \ref{section:Intro}, to select an XAI solution, the data scientist can currently rely on XAI libraries, benchmarks, and AutoML frameworks. 
Currently, available XAI libraries such as DeepExplain \cite{ancona2017towards}, AI Explainability 360 \cite{aix360-sept-2019}, and Alibi \cite{JMLR:v22:21-0017} are gathering state-of-the-art XAI solutions. However, they neither integrate automatic evaluation of explanation nor recommend XAI solutions according to data scientists' needs and constraints. 

Comparatives and benchmarks \cite{yeh2019fidelity, hooker2019benchmark, alvarez2018robustness} compare XAI solutions efficiency using XAI evaluation metrics. They are often joint with the proposal of an XAI solution or an XAI evaluation metric on which they are focusing. However, the results obtained depend on the dataset and the ML model that may not be the ones the data scientist uses, and thus the results may be different.
Moreover, the hyperparameters of the XAI solutions are not optimized to maximize the various properties needed by the data scientist. This last point is problematic as some XAI properties such as correctness and compactness are not independent \cite{nauta2022anecdotal}.

Eventually, \cite{Axa} highlights that users should be guided in choosing XAI solutions and proposes a methodology for this issue, while \cite{Palacio_2021_ICCV} proposes a theoretical framework to facilitate the comparison between XAI solutions.\\

To summarize, as there is a high diversity of XAI solutions, it is a complex and tedious task for data scientists to find XAI solutions that fit their needs. Yet, there is no recommendation system to automate this task. Moreover, data scientists look for the best XAI solution as they want a reliable solution. XAI evaluation metrics allow for objective comparison, but XAI libraries do not implement them and comparatives are not adapted to the user's context. Eventually, data scientists have to find hyperparameters to maximize the desired properties. However, this task should be done for multiple XAI solutions and multiple properties according to the data scientists' preferences. This paper aims to address these issues in the following sections.

\section{Example and formalization}\label{section:exandformal}
\subsection{Illustrative example}
\label{section:scenario}

Let's first consider a data scientist in a medical laboratory, Alice and Bob, a physician colleague. Bob uses a ML black box model as a decision support tool and asks if it is possible to have an explanation for the predictions of the model to check some rare cases. Alice has access to the model, as well as the data that were used to build it, and now wants to implement an XAI solution to produce explanations.

Here, the needs of Bob, the physician, are the following: the explanations must focus on predictions (as it is asked why they are obtained) and the XAI solution must explain a trained model without modifying it. Moreover, Bob wants to know how the collected data for a patient (the features) influence the result of the model.

Regarding the constraints of the context, the high-stakes decisions impose the use of a precise model and the most faithful explanations possible (correctness property). Nevertheless, the explanations should not be completely changed by small perturbations as blood measurements might be noisy, therefore, stable explanations are mandatory (continuity property). Eventually, since Bob will be the main user of these explanations, concise explanations should be encouraged as physicians shouldn't waste time on unimportant features (compactness property).

\subsection{Definitions}\label{section:formalization}

\begin{definition}[Dataset]
Let $X,Y$ be a dataset with $X = \{x_i\}_{i=1}^n | x_i \in \mathbb{R}^d$ the observations and $Y = \{y_i\}_{i=1}^n | y_i \in \mathbb{R}$ their corresponding labels, $n$ is the number of observations and $d$ is the number of dimensions of the dataset (also called features). 
\end{definition}

\textit{In our illustrative example}, the diabetes dataset \cite{efron2004least} $X$ is composed of $n=442$ patients with $d=10$ features (physiological information) and $Y$ is the disease progression for each patient.

\begin{definition}[ML model]
A ML model is trained on a dataset $X,Y$ by inferring statistical relationships between $X$ and $Y$. This model can then be used as predictive function which we note $f:X \rightarrow \hat{Y}$ with $\hat{Y} = \{\hat{y}_i\}_{i=1}^n | \hat{y}_i \in \mathbb{R}$, the produced predictions.
\end{definition}

\textit{In our illustrative example}, a Multilayer Perceptron learns a predictive function $f$ that predicts the disease progressions $\hat{Y}$ based on the observations $X$ and the labels $Y$.

\begin{definition}[Explanandum and explanan]
We note $\mathcal{E} = \{\mathcal{E}_i\}_{i=1}^k$ the set of all possible explanandum, where the explanandum $\mathcal{E}_i$ is a descriptor for explanation functions that specifies \textit{what is explained}. We also note $\mathcal{E}^\prime = \{\mathcal{E}^\prime_j\}_{j=1}^{k^\prime}$ the set of all possible explanan, where the explanan $\mathcal{E}^\prime_j$ is a descriptor for explanation functions that specifies \textit{how it is explained}.
\end{definition}

\textit{In our illustrative example}, $\mathcal{E}_i=$\textit{Why this prediction?} and $\mathcal{E}^\prime_j=$\textit{feature summaries}.

\begin{definition}[XAI properties]
XAI properties are descriptive quality criteria for explanations. We note $P_r$, the set properties that explanations verify or not.
\end{definition}

\textit{In our illustrative example}, the properties of interest are correctness, continuity, and compactness.

Thus, in AutoXAI the data scientist can specify the needs with $(\mathcal{E},\mathcal{E}^\prime)$ and constraints with $P_r$.

\begin{definition}[XAI solution]
An XAI solution acts as a function that produces one or several explanations. We note $E =\{e_t\}_{t=1}^l$, the explanations set with $l \in \mathbb{N}$ the number of explanations. We note $f_e^{(h)} : P(X,Y,F,\hat{Y}) \to E$ the explanation function with $P(X,Y,F,\hat{Y})$ a partition of $\{X,Y,F,\hat{Y}\}$ and $h$ the hyperparameters of the XAI solution. $f_e^{(h)} \in F_e$ with $F_e$ the set of explanation functions. The hyperparameters refer to the static parameters that determine the behaviors of the XAI solution. For naturally interpretable models $f=f_e^{(h)}$. 
\end{definition}

\textit{In our illustrative example}, the XAI solutions LIME \cite{ribeiro2016should} and Kernel SHAP \cite{lundberg2017unified} are considered as $f_e^{(h)} : (X,F) \to E$. Both produce one feature importance explanation $e_t \in \mathbb{R}^d$ for each patient so that $E =\{e_t\}_{t=1}^n$.

\begin{definition}[XAI evaluation metrics]\label{def:metric}

An XAI evaluation metric evaluates one property and is often adapted to one specific type of explanation. We note the set of XAI evaluation metrics $M = \{m_q\}_{q=1}^{c}$, where $m_q : P(X,F,F_e,Y) \to \mathbb{R}$, with $P(X,F,F_e,Y)$ a partition of $\{X,F,F_e,Y\}$, so that $m_q$ evaluates $p_q\in P_r$.
\end{definition}
\textit{In our illustrative example}, robustness $m_q:(X,F_e)\to \mathbb{R}$ is an evaluation metric that assess the property of continuity $p_q$.

\section{Our framework proposal}\label{section:framework}

\begin{figure*}[t!]
\centering

\begin{subfigure}{\textwidth}
\begin{tikzpicture}[every text node part/.style={align=center}]

\fill (0,0)ellipse(0.12 and 0.25) (0,0.35)circle(0.1);

\node[font=\large] (user0) at (0,-1){User\\(1)};
\node[font=\large] (user0) at (5,-1){\\(2)};
\node[font=\large] (user0) at (12,-1){\\(3)};
\node[font=\large] (user) at (0,0){};
\node[draw, rounded corners=2pt, font=\large] (ihm) at (5,0){Context\\Adapter};
\node[draw, rounded corners=2pt, font=\large] (hpo) at (12,0){Hyperparameters\\Optimizer};

\draw[-latex] (user) -- node [midway, above, font=\small] {Context : needs $(\mathcal{E}_i,\mathcal{E}^\prime_j)$,\\ constraints $\subset P_{r\mathcal{E}_i,\mathcal{E}^\prime_j}$,\\ data $(X,Y)$, model $f$,\\ parameters for AutoXAI,\\ preferences for properties} (ihm);

\path (ihm.east) -- (ihm.north east) coordinate[pos=0.5] (a1);
\path (hpo.west) -- (hpo.north west) coordinate[pos=0.5] (b1);
\draw[-latex] (a1) to[bend left=40] node [midway , above, font=\small] {XAI solutions $\subset F_{e\mathcal{E}_i,\mathcal{E}^\prime_j}$ \\Evaluation metrics $\subset M_{\mathcal{E}_i,\mathcal{E}^\prime_j}$ \\\\ $f_{e1}$} (b1);

\draw[-latex] (ihm.east) to node [midway , above, font=\small] {$f_{e2}$} (hpo.west);

\path (ihm.east) -- (ihm.north east) coordinate[pos=-0.5] (a1);
\path (hpo.west) -- (hpo.north west) coordinate[pos=-0.5] (b1);
\draw[-latex] (a1) to[bend right=40] node [midway , above, font=\small] {$f_{e3}$} (b1);
\draw[-latex] (hpo) to[bend left=50] node [midway, below, font=\small] {XAI solutions optimized and ranked} (user);

\end{tikzpicture}
\caption{Global architecture of AutoXAI}
\label{fig:archia}
\end{subfigure}
~
\begin{subfigure}{0.6\textwidth}
\begin{tikzpicture}[every text node part/.style={align=center}]

\node[] (input) at (-4,0){};
\node[] (output) at (-4,-4){};
\node[] (hpel) at (-0.5,0.75){(3.1)};
\node[] (exl) at (6.5,0.5){(3.2)};
\node[] (evl) at (3,-3.5){(3.3)};
\node[draw, rounded corners=2pt, font=\large] (hpe) at (0,0){Hyperparameter\\Estimator};
\node[draw, rounded corners=2pt, font=\large] (ev) at (3,-3){Evaluator};
\node[draw, rounded corners=2pt, font=\large] (ex) at (6,0){Explainer};

\draw[-latex] (input) -- (hpe) node [near start, above, font=\small] {XAI solution and \\evaluation metrics};
\draw[-latex] (hpe) to[bend left=80] node [midway , above, font=\small] {New hyperparameters} (ex);
\draw[-latex] (ex) to[bend left=40] node [midway, sloped, below, font=\small] {Explanations} (ev);
\draw[-latex] (ev) to[bend left=40] node [midway, sloped , below, font=\small] {Aggregated score} (hpe);
\draw[-] (ev) to[bend left=10] node [near end, above, font=\small] {XAI solutions optimized \\and ranked} (output);

% \fill (0,0)ellipse(0.12 and 0.25) (0,0.35)circle(0.1);

% \node[font=\large] (user0) at (0,-1){User\\(1)};
% \node[font=\large] (user0) at (5,-1){\\(2)};
% \node[font=\large] (user0) at (12,-1){\\(3)};
% \node[font=\large] (user) at (0,0){};
% \node[draw, rounded corners=2pt, font=\large] (ihm) at (5,0){Context\\Adapter};
% \node[draw, rounded corners=2pt, font=\large] (hpo) at (12,0){Hyperparameters\\Optimizer};

% \draw[-stealth ] (user) -- node [midway, above, font=\small] {Context : needs $(\mathcal{E}_i,\mathcal{E}^\prime_j)$,\\ constraints $\subset P_{r\mathcal{E}_i,\mathcal{E}^\prime_j}$,\\ data $(X,Y)$, model $f$, ...} (ihm);

% \path (ihm.east) -- (ihm.north east) coordinate[pos=0.5] (a1);
% \path (hpo.west) -- (hpo.north west) coordinate[pos=0.5] (b1);
% \draw[-stealth] (a1) to[bend left=40] node [midway , above, font=\small] {XAI solutions $\subset F_{e\mathcal{E}_i,\mathcal{E}^\prime_j}$ \\Evaluation metrics $\subset M_{\mathcal{E}_i,\mathcal{E}^\prime_j}$ \\\\ $f_{e1}$} (b1);

% \draw[-stealth] (ihm.east) to node [midway , above, font=\small] {$f_{e2}$} (hpo.west);

% \path (ihm.east) -- (ihm.north east) coordinate[pos=-0.5] (a1);
% \path (hpo.west) -- (hpo.north west) coordinate[pos=-0.5] (b1);
% \draw[-stealth] (a1) to[bend right=40] node [midway , above, font=\small] {$f_{e3}$} (b1);
% \draw[-stealth] (hpo) to[bend left=50] node [midway, below, font=\small] {XAI solutions optimized and ranked} (user);

\end{tikzpicture}
\caption{Details on Hyperparameters Optimizer}
\label{fig:archib}
\end{subfigure}

\caption{Architecture of AutoXAI.}
The figures are read by following the number of the steps. In Figure \ref{fig:archia}, for each XAI solution, step (3) optimizes its hyperparameters with respect to the aggregated scores of the evaluation metrics by entering the loop in Figure \ref{fig:archib}.

\label{fig:archi}
\end{figure*}
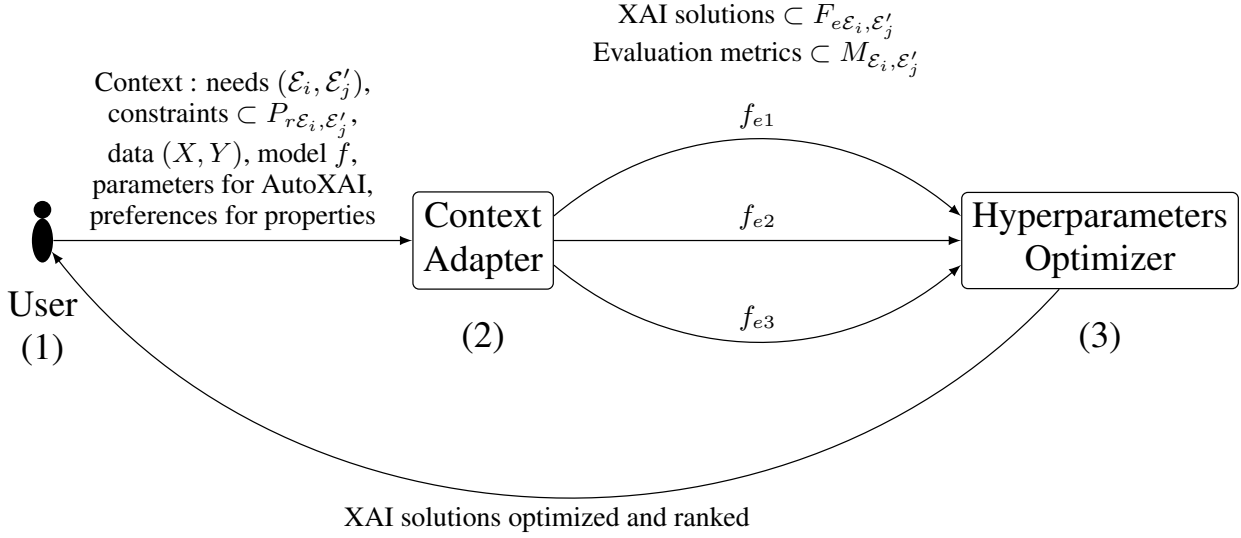
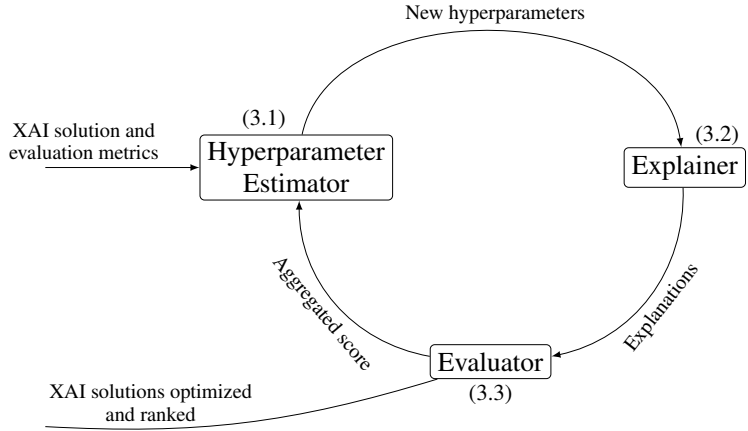

We first describe a global step-by-step AutoXAI process with its components as shown in Figure \ref{fig:archia}, then we detail the Hyperparameters Optimizer in Figure \ref{fig:archib} and eventually each component using the definitions of Section \ref{section:formalization}.

Here are the operations as they are performed, starting with Figure \ref{fig:archia}:

\begin{enumerate}
    \item The User gives the elements of the context, the parameters for AutoXAI and its preferences regarding the properties.
    \item The Context Adapter component selects a subset of XAI solutions matching the needs and a subset of evaluation metrics to ensure that the constraints are met.
    \item For each XAI solution, the Hyperparameters Optimizer looks for hyperparameters that will reduce the loss function based on the aggregated scores of the evaluation metrics. To do so, it performs the following operations in a loop, see Figure \ref{fig:archib}.
    \begin{enumerate}
        \item The Hyperparameters Estimator proposes new hyperparameters according to the chosen optimization algorithm.
        \item The Explainer uses the XAI solution and the newly proposed hyperparameters to produce explanations.
        \item The Evaluator applies the evaluation metrics to the explanations and aggregates the scores thus obtained.
    \end{enumerate}

\end{enumerate}

\subsection{Context adapter}\label{subsection:context}

As detailed in Section \ref{section:formalization}, XAI solutions can be grouped according to $(\mathcal{E}_i,\mathcal{E}^\prime_j)$, their explanandum and their explanan. This grouping also determines $P_{r\mathcal{E}_i,\mathcal{E}^\prime_j}$, the properties that can describe the XAI solutions, and therefore $M_{\mathcal{E}_i,\mathcal{E}^\prime_j}$, the XAI evaluation metrics that can be applied.
To get $(\mathcal{E}_i,\mathcal{E}^\prime_j)$, we ask what the user wants in natural language with pre-written answers. To do so, AutoXAI uses the question bank from \cite{liao2020questioning} for explanandum proposals, letting the user choose which question the XAI solution should answer. For the explanan, AutoXAI uses the list of explanation types from \cite{carvalho2019machine}.
With this knowledge, a contextual prefiltering (see Section \ref{subsection:CARS}) is possible by selecting $F_{e\mathcal{E}_i,\mathcal{E}^\prime_j}$ in $F_e$, $P_{r\mathcal{E}_i,\mathcal{E}^\prime_j}$ in $P_r$ and $M_{\mathcal{E}_i,\mathcal{E}^\prime_j}$ in $M$. This is possible by tagging each of these proposals with a tuple from $(\mathcal{E},\mathcal{E}^\prime)$ using \cite{liao2020questioning,nauta2022anecdotal} correspondence tables.
$P_{r\mathcal{E}_i,\mathcal{E}^\prime_j}$ serve for contextual modeling (see Section \ref{subsection:CARS}). Indeed, the user can choose the weights (degree of importance) for each of the properties in $P_{r\mathcal{E}_i,\mathcal{E}^\prime_j}$. These weights are used in the evaluation of the XAI solutions which guide the optimization and therefore the ranking of XAI solutions that are produced. 1 is the default value, increasing a weight $w_q$ makes its property $p_q$ more important and conversely, 0 means that the property is ignored. In addition to these elements, AutoXAI can retrieve the model and the dataset on which it was trained if necessary.

\subsection{Hyperparameters estimator}

This component's goal is to propose new hyperparameters to obtain the best aggregated score. The optimization algorithms are iterative and some, such as Bayesian optimization \cite{nogueira2014bayesian}, associate the hyperparameters to a score for building a probabilistic model. This probabilistic model estimates the hyperparameters that should give the best score. Here, $f_e$, with previously estimated hyperparameters, is evaluated with the aggregated scores of $M_{\mathcal{E}_i,\mathcal{E}^\prime_j}$ (further detailed in Section \ref{subsection:evaluator}) and it results in a new entry to update the probabilistic model. This component estimates hyperparameters according to previous score results in case there are. Otherwise, the values of the hyperparameters are set according to the initialization of the chosen algorithm, for example at random.

\subsection{Explainer component}
This component's objective is to produce explanations using the XAI solution and the defined hyperparameters.
As the implementations of XAI solutions vary in their programming paradigm and in the data structure they use and return, it is necessary to set up a wrapper that standardizes the input and output for each XAI solution of a given group. This component serves as a base to include all implemented XAI solutions and is made to be completed with new XAI solutions.

\subsection{Evaluator component}\label{subsection:evaluator}
This component aims at computing the scores of XAI evaluation metrics corresponding to the XAI properties requested by the user. Like the precedent component, this one also acts as a wrapper that standardizes the input and output of XAI evaluation metrics and serves as a base to include any XAI evaluation metrics.

It also aggregates the properties' scores to provide a unique optimization objective for HPO. To find the best hyperparameters $h$ of the XAI solution $f_e$, we define the optimization objective as follows: 
\begin{equation}
\max_{h\in H}A(f_e,h)
\end{equation}
With $A$ the aggregation function and $H$ the set of every possible hyperparameters.
$A$ should gather the multiple scores returned by the XAI evaluation metrics assessing the properties chosen and weighted according to the user's preferences for properties. To do so, we opt for a linear scalarization \cite{miettinen2012nonlinear}:
\begin{equation}\label{eq:agg_score}
A(f_e,h)=\frac{1}{c^\prime}\sum _{q=1}^{c^\prime}w_{q} \times sc_q(m_q(f_e^{(h)}))
\end{equation}
The XAI evaluation metric for $p_q$ is written $m_q(f_e^{(h)})$ for a shorter notation, thought it might use any partition of $\{X,F,F_e,Y\}$ as defined in \ref{def:metric}.  $c^\prime$ is the number of chosen properties and therefore XAI evaluation metrics. The weights $w_q$ are the degree of importance set by the user and $sc_q(.)$ is a scaling function based on previous results for a property $p_q$. This scaling function allows not to favor one XAI evaluation metric over another. For the first epoch, there are no previous results and scaling cannot be defined. Therefore, we initialize $sc_q(.)$ using the evaluation scores of XAI solutions with the default hyperparameters. This cold start also verifies whether XAI solutions can perform well with default hyperparameters.

Although this framework already produces a ranking of XAI solutions, the computation time should be taken into account. 

\section{Time-saving evaluation strategies}\label{section:time}

Some XAI solutions and XAI evaluation metrics were not designed to be used multiple times in a row and have high algorithmic complexity. To reduce the time cost of these algorithms, without changing their architecture, we propose to adapt the existing heuristic strategies from the AutoML field. AutoXAI adapts three AutoML strategies to reduce computing time: low fidelity which becomes sampling, early stopping, and weight sharing which becomes information sharing. These strategies are detailed below.

\subsubsection{Sampling}
Reducing the number of explanations produced reduces the number of operations in explanation making and in explanation evaluation. It is also possible to use a subset of the dataset to create explanations, which can reduce the number of operations for explanation making.
These approximations can be accurate enough depending on the diversity of the dataset, the complexity of the model, and the sensitivity of the XAI solution. This issue is addressed by specifying the percentage of explanations to be processed.

\subsubsection{Early stopping}
Early stopping can be performed during the calculation of the XAI evaluation metric or the HPO. In both cases, this strategy can save a lot of time but it must be set up correctly to avoid approximating too roughly. For the calculation of evaluation metrics, this option is only possible if the XAI evaluation metric is applied sequentially to the explanations. Moreover, this strategy is more efficient if explanations are also calculated sequentially, indeed, fewer explanations are computed this way. The stopping condition is that the XAI evaluation metric only updates itself by a small percentage below a threshold during multiple iterations. We then consider that it stabilizes.
For early stopping applied to HPO, the reasoning is that if the best score does not change during several iterations, then it has been found. Here, the choice of a threshold matters to avoid missing a better solution.

\subsubsection{Information sharing}
Another way to save computing time is to reuse intermediate results and share information between evaluations. In AutoML, the weight-sharing strategy of an old model speeds up the training of a new one \cite{he2021automl}. In AutoXAI, some intermediate calculations can be reused in the same way. This strategy is especially efficient for some costly intermediate calculations like a Gaussian process \cite{alvarez2018robustness}.

\begin{table*}[]
\caption{XAI evaluation metrics}
\resizebox{\textwidth}{!}{
\begin{tabular}{|c|c|c|}
\hline
\textbf{Property}&\textbf{Metric}&\textbf{Formula}\\ \hline
\textit{Continuity}&\textbf{Robustness}&$\hat{L}(x_i)=\max\limits_{x_j\in B_\epsilon(x_i)}\frac{\|f_e^{(h)}(x_i)-f_e^{(h)}(x_j)\|_2}{\|x_i-x_j\|_2}$ \\&& \parbox{17cm}{With $B_\epsilon(x_i)$ a ball of radius $\epsilon$* centered in $x_i$ the studied data point.
The optimization method looks for the point
in the data space with the highest ratio in the neighborhood.
This value is kept as a measure for the lack of robustness $\hat{L}(x_i)$.}\\ \hline
\textit{Correctness}&\textbf{Infidelity}&$INFD(e_i,f,x_i) = \mathbb{E}_{I\sim\mu I}[(I^Te_i-(f(x_i)-f(x_i-I)))^2]$ \\&& With $I$, perturbations around $x_i$ so that $I = x_i - x_i'$, here we choose a noisy implementation with $x_i' = x_i + \epsilon$ and $\epsilon$ a uniform noise.\\ \hline
\textit{Compactness}&\textbf{Number of features}&$NoF(e_i) = \operatorname{Card}(e_i)$ \\&& With $e_i \in E$, the $i$-th explanation vector, a feature summary.\\ \hline
\textit{Completeness}&\textbf{Non-representativeness}& $NR(E) = \frac{1}{n}\sum\limits_{i=1}^n \min\limits_{e_j\in E} d(x_i,e_j)$ \\&& With $n=\operatorname{Card}(X)$, $d$ a distance function and $E$ the explanations set composed of representative data points $e_j$ also called prototypes.\\ \hline
\textit{Compactness}&\textbf{Diversity}&$Div(E) = \sum\limits_{\{e_i,e_j\} \in P_2(E)}\frac{d(e_i,e_j)}{C_2^l} $\\(\textit{Redundancy})&& With $P_2(E)$, the set of combinations of two prototypes, $d$ a distance function and $C_2^l$ the number of combination with $l=\operatorname{Card}(E)$ prototypes.\\ \hline
\textit{Compactness}&\textbf{Number of prototypes}& $NoP(E) = \operatorname{Card}(E)$ \\(\textit{Size})&& With $E$, the prototypes set. \\ \hline
\multicolumn{3}{l}{\small{*The implementation is a box having the size of the standard deviation, it is a variation proposed by the original authors.}}
\end{tabular}}
\label{tab:evaluation}
\end{table*}

\section{Experiments}\label{section:experiments}
In the following experiments, an implementation of the AutoXAI framework described in Section \ref{section:framework} is applied to two use cases including the illustrative example described in Section \ref{section:scenario}. The code to reproduce the results of these use cases is available at \url{https://github.com/RobinCugny/AutoXAI}. 

\subsection{Diabetes estimation}\label{subsection:expe_diab}
For the illustrative example, the datasets used are diabetes dataset \cite{efron2004least} and Pima Indians dataset \cite{smith1988using}. Diabetes dataset has 10 features and is designed for a regression task to predict disease progression. Pima Indians dataset has 8 features and is made for a binary classification to predict if patients have diabetes.
The black box model used is the scikit-learn implementation of a Multilayer Perceptron \cite{scikit-learn}. For the regression we use MLPRegressor\footnote{\url{https://scikit-learn.org/stable/modules/generated/sklearn.neural_network.MLPRegressor.html}} and for the classification we use MLPClassifier\footnote{\url{https://scikit-learn.org/stable/modules/generated/sklearn.neural_network.MLPClassifier.html}}.

The implemented XAI solutions are LIME \cite{ribeiro2016should} and Kernel SHAP \cite{lundberg2017unified}. 
LIME provides explanations using the weights of a local linear model for each observation.
SHAP captures features' interactions using Shapley values \cite{shapley1953} as previously done by \cite{strumbelj2010efficient}. 
We use Kernel SHAP, one of their contributions, that builds a linear model like LIME but uses Shapley values as coefficients and therefore as feature importance. In this paper, SHAP refers to Kernel SHAP for short.

The implemented XAI evaluation metrics and their corresponding properties are:
\begin{itemize}
    \item \textbf{Robustness} \cite{alvarez2018robustness} \textit{Continuity}
    \item \textbf{Infidelity} \cite{yeh2019fidelity} \textit{Correctness}
    \item \textbf{Number of features}  \cite{rosenfeld2021better} \textit{Compactness}
\end{itemize}

\cite{alvarez2018robustness} proposes the measure of \textbf{robustness} that evaluates \textit{Continuity} by adapting Lipschitzian continuity. The objective is to measure changes in explanations while the input has small perturbations. Indeed, explanations should not radically change if the observation does not either. \cite{yeh2019fidelity} proposes to evaluate \textit{Correctness} with an \textbf{infidelity} metric. It consists in perturbing the input, based on the order of the features in explanations, and measuring for each new input the change in the output of the predictive function $f$. The \textit{Compactness} is evaluated with the \textbf{number of features} which is obtained with the cardinal of the explanation vector. See Table \ref{tab:evaluation} for the formulas.

For the aggregation in this scenario, Alice and Bob set the weights to 1, 2, 0.5 for \textbf{robustness}, \textbf{infidelity} and \textbf{number of features} respectively. Alice, sets the number of epochs to 25.

The HPO strategy is a Bayesian optimization, we use \cite{nogueira2014bayesian} implementation of the Gaussian Process.
The time-saving evaluation strategies are early stopping for XAI evaluation metrics computation and HPO, and information sharing for robustness and infidelity. For robustness, information shared between epochs are the data points giving the maximum score of robustness (see Table \ref{tab:evaluation}) we call maxima for short. For infidelity, the information shared between epochs are the generated perturbation points and the model predictions for them.

An extract from the ranking produced by AutoXAI for Diabetes dataset is in Table \ref{table:results_diab} and the one for Pima Indians dataset is in Table \ref{table:results_pima}. The XAI solutions are sorted in descending order according to the Aggregated score produced by Equation \ref{eq:agg_score}.
To show diverse XAI solutions we present three combinations of hyperparameters with LIME and three with SHAP. The choice of the number of features in the explanation is a subjective process requiring to visualize the explanations, we decide on a number of features of 1, 3, and 5. Thus, the user can verify if short explanations are enough to understand the prediction or if more features would help. In the Hyperparameter columns, the two first hyperparameters for LIME as for SHAP are: first, the number of features in the explanation, and second, the number of perturbations used to build the linear model. The last hyperparameter for SHAP is the l1 regularization to use for feature selection\footnote{\url{https://shap-lrjball.readthedocs.io/en/latest/generated/shap.KernelExplainer.html}}.

\begin{table*}[]
\centering
\small
\caption{Extract from the ranking produced by AutoXAI on Diabetes dataset.}
\begin{tabular}{@{}c|ccc|cc@{}}

\hline
\begin{tabular}[c]{@{}l@{}}Aggregated\\score\end{tabular}  &
\begin{tabular}[c]{@{}l@{}}Scaled\\Robustness\end{tabular} & 
\begin{tabular}[c]{@{}l@{}}Scaled\\Fidelity\end{tabular}& 
\begin{tabular}[c]{@{}l@{}}Scaled\\NoF\end{tabular}& 
\begin{tabular}[c]{@{}l@{}}XAI\\Solution\end{tabular}& 
Hyperparameters 
\\ \hline
1.023 & 0.727 & 0.833 & 1.351 & LIME & 1;3656 \\
1.019 & 0.703 & 0.991 & 0.745 & LIME & 3;8782 \\
0.963 & 0.682 & 1.068 & 0.139 & LIME & 5;5392 \\
-0.287 & 0.310 & -0.924 & 1.351 & SHAP & 1;1304;auto \\
-0.633 & -0.319 & -0.975 & 0.745 & SHAP & 3;1571;aic \\
-0.639 & 0.014 & -1.000 & 0.139 & SHAP & 5;1148;aic
\end{tabular}
\label{table:results_diab}
\end{table*}

\begin{table*}[]
\centering
\small
\caption{Extract from the ranking produced by AutoXAI on Pima Indians dataset.}
\begin{tabular}{@{}c|ccc|cc@{}}
\hline
\begin{tabular}[c]{@{}l@{}}Aggregated\\score\end{tabular}  &
\begin{tabular}[c]{@{}l@{}}Scaled\\Robustness\end{tabular} & 
\begin{tabular}[c]{@{}l@{}}Scaled\\Fidelity\end{tabular}& 
\begin{tabular}[c]{@{}l@{}}Scaled\\NoF\end{tabular}& 
\begin{tabular}[c]{@{}l@{}}XAI\\Solution\end{tabular}& 
Hyperparameters \\ \hline
1.412 & 0.744 & 1.435 & 1.243 & LIME & 1;5347 \\
1.282 & 0.575 & 1.325 & 1.243 & SHAP & 1;509;bic \\
0.361 & 0.633 & 0.117 & 0.430 & LIME & 3;8329 \\
0.176 & 0.339 & -0.014 & 0.430 & SHAP & 3;713;auto \\
0.070 & 0.262 & 0.070 & -0.383 & SHAP & 5;537;bic \\
-0.185 & 0.599 & -0.481 & -0.383 & LIME & 5;7023  
\end{tabular}

\label{table:results_pima}
\end{table*}

\begin{figure}
    \begin{subfigure}{.5\textwidth}
        \includegraphics [width=\textwidth]{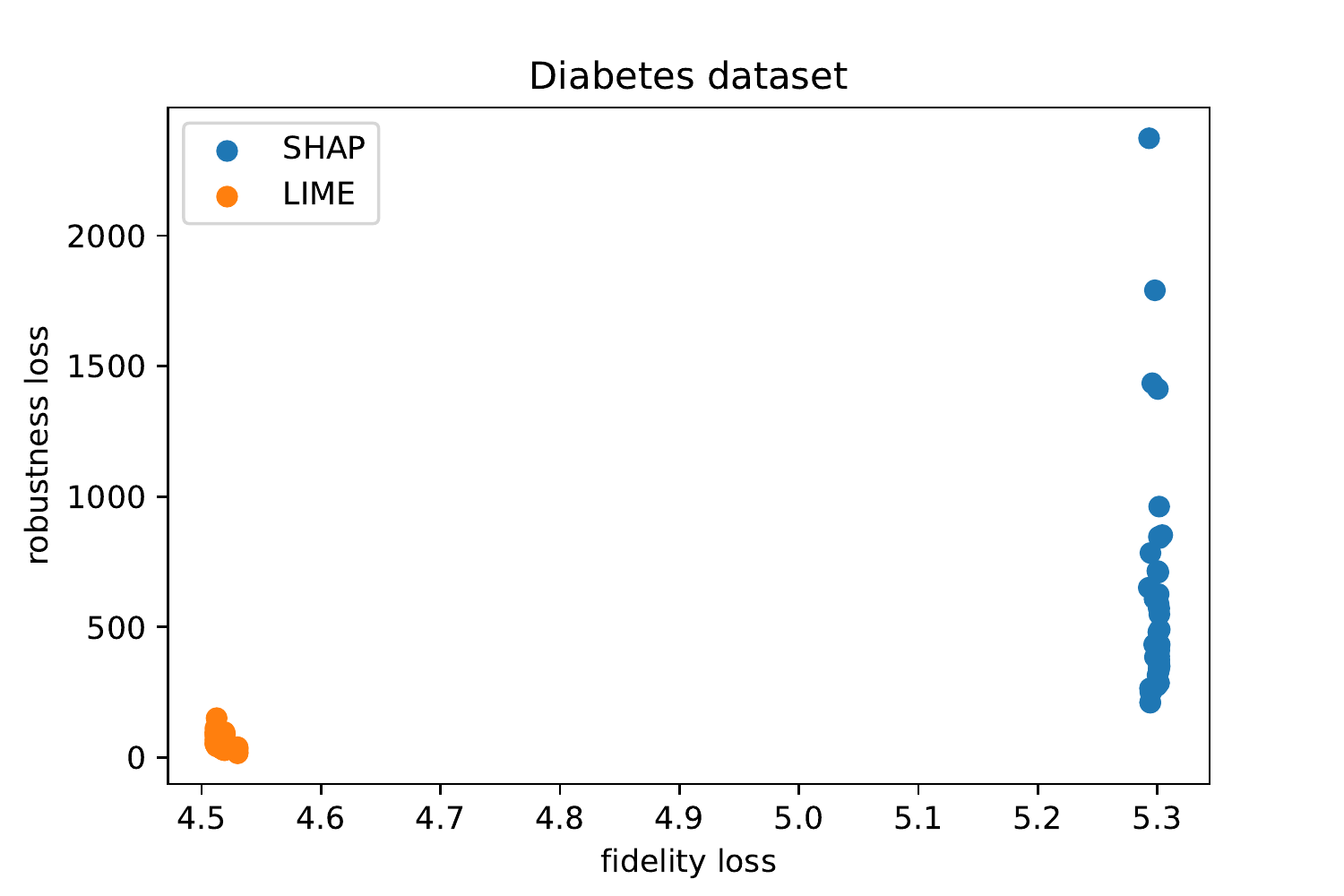}
        \caption{Scores of XAI solutions for Diabetes dataset}
        \label{fig:diab_scores}
    \end{subfigure}
   ~ 
    \begin{subfigure}{.5\textwidth}
        \includegraphics [width=\textwidth]{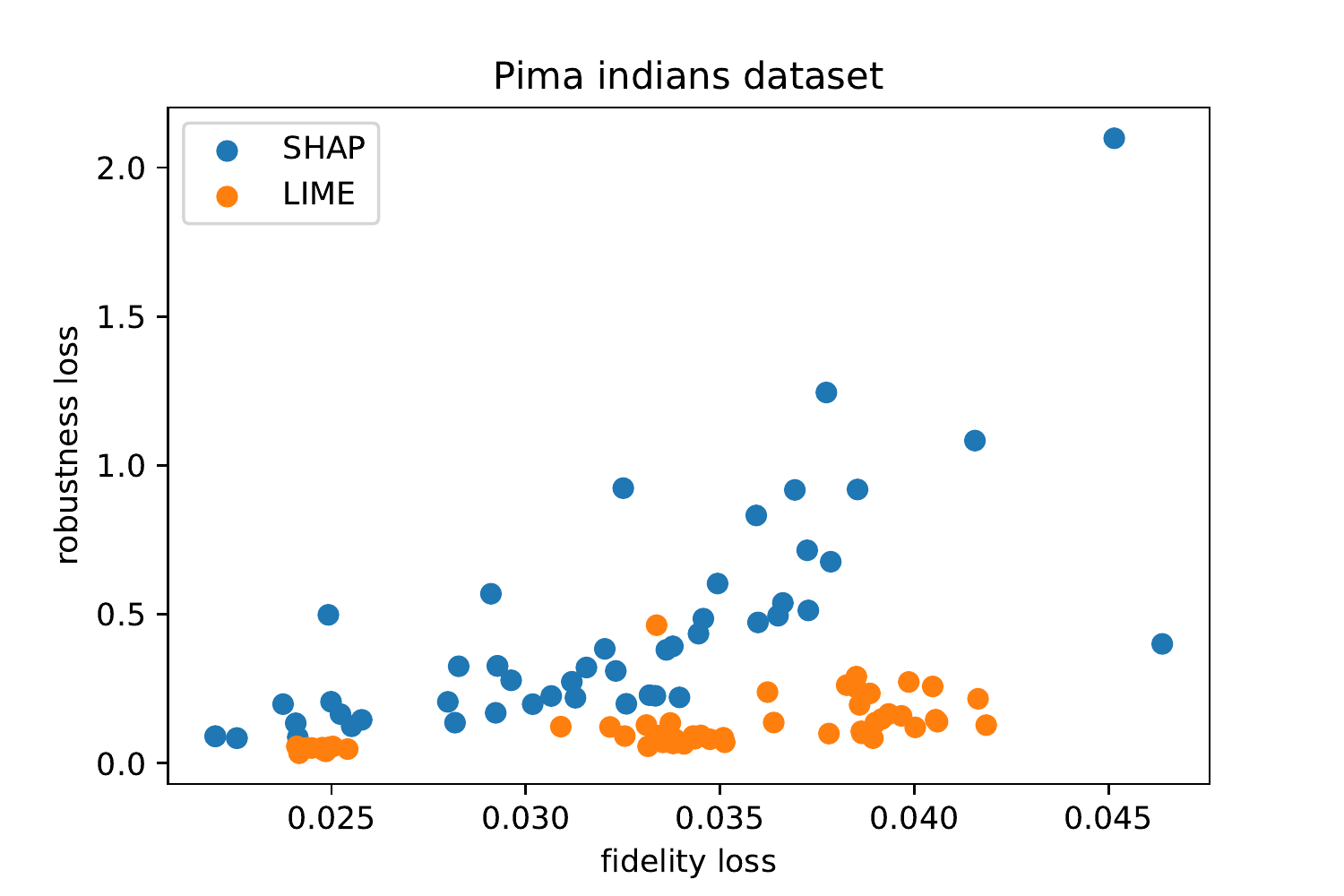}
        \caption{Scores of XAI solutions for Pima indians dataset}
        \label{fig:pima_scores}
    \end{subfigure}
    \caption{Robustness loss and Fidelity loss}
\end{figure}

Regarding Table \ref{table:results_diab}, LIME is systematically higher than SHAP in the rankings with these XAI evaluation metrics. Several factors could explain why SHAP does not score better: with default hyperparameters, SHAP has lower average robustness than LIME on certain UCI classification datasets \cite{Dua:2019} (glass, wine, and leukemia) and especially a larger standard deviation \cite{alvarez2018robustness}. For Diabetes dataset, as we can see in Figure \ref{fig:diab_scores}, we also observe a large standard deviation and higher average robustness loss for SHAP with different hyperparameters. We observe especially that it has constantly a higher fidelity loss with a narrow distribution on this score. According to \cite{yeh2019fidelity}, this means that SHAP captures less well how the model prediction changes in response to the perturbations. As Kernel SHAP relies on LIME strategy to build a local linear model, this means that the loss might come from the inclusion of Shapley values. A hypothesis here would be that the black box model does not make much use of the relationships between features or not in the same way as Kernel SHAP detects it with its linear approximation. Besides, a similar conclusion has already been observed in \cite{DBLP:conf/dolap/DoumardAEEMS22}.

Regarding the Pima Indians dataset, however, it seems that SHAP performs slightly better. In Figure \ref{fig:pima_scores}, we can see that SHAP is less robust than LIME most of the time but that it has equivalent fidelity. Here SHAP seems to succeed in capturing the predictor function changes, therefore it might find more feature interactions in common with the model.

\begin{figure}
\includegraphics [width=0.49\columnwidth]{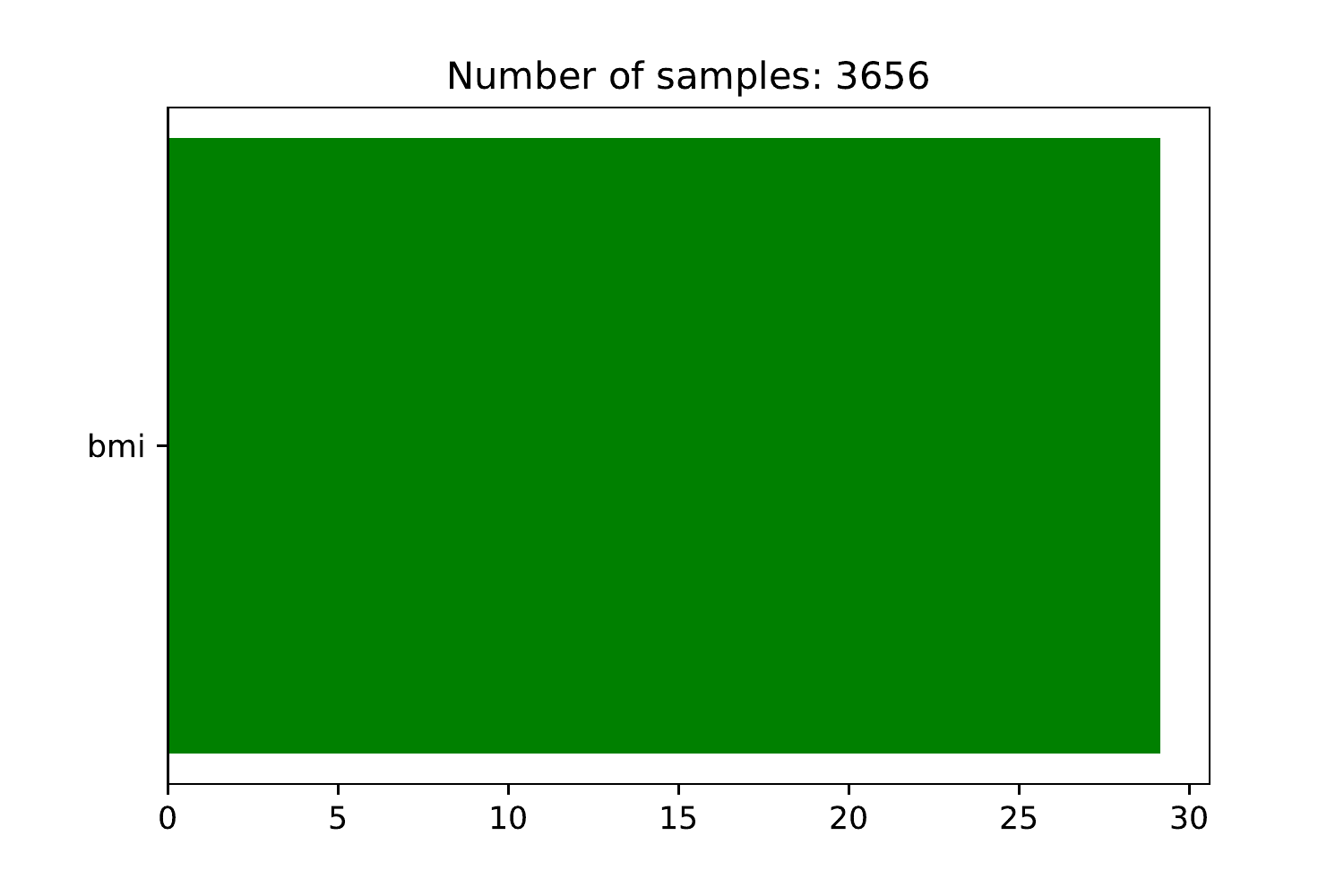}
\includegraphics [width=0.49\columnwidth]{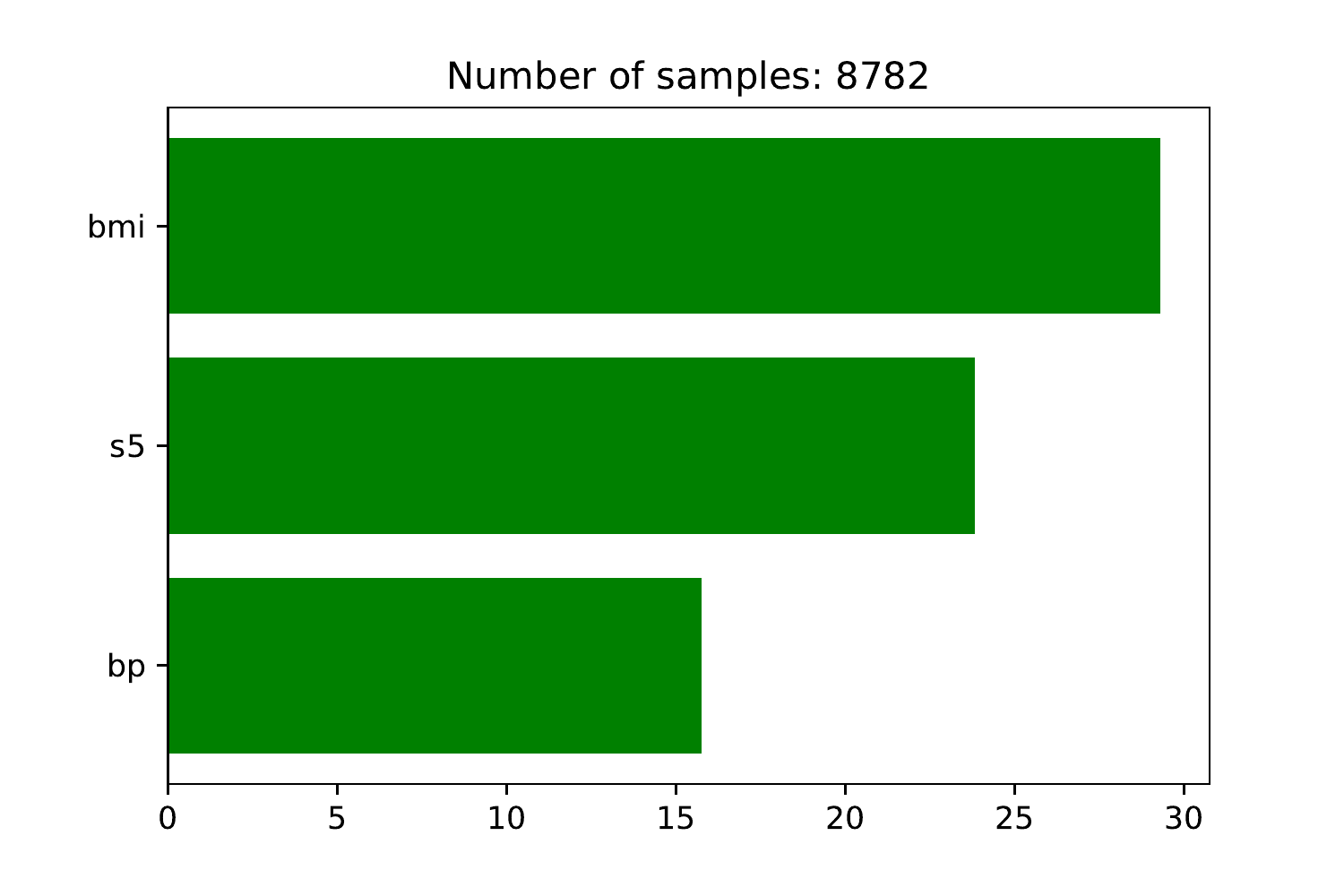}
\includegraphics [width=0.49\columnwidth]{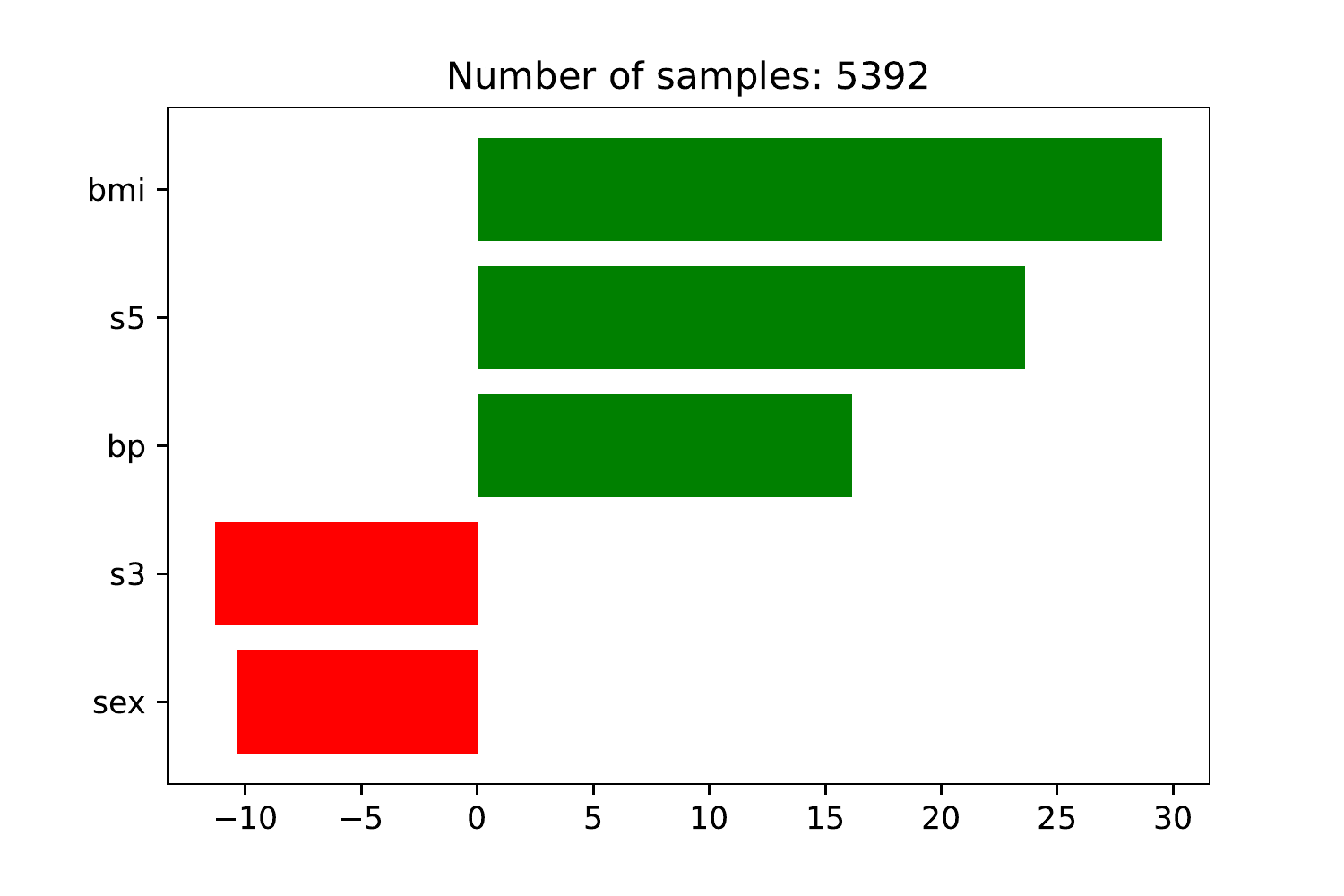}
\includegraphics [width=0.49\columnwidth]{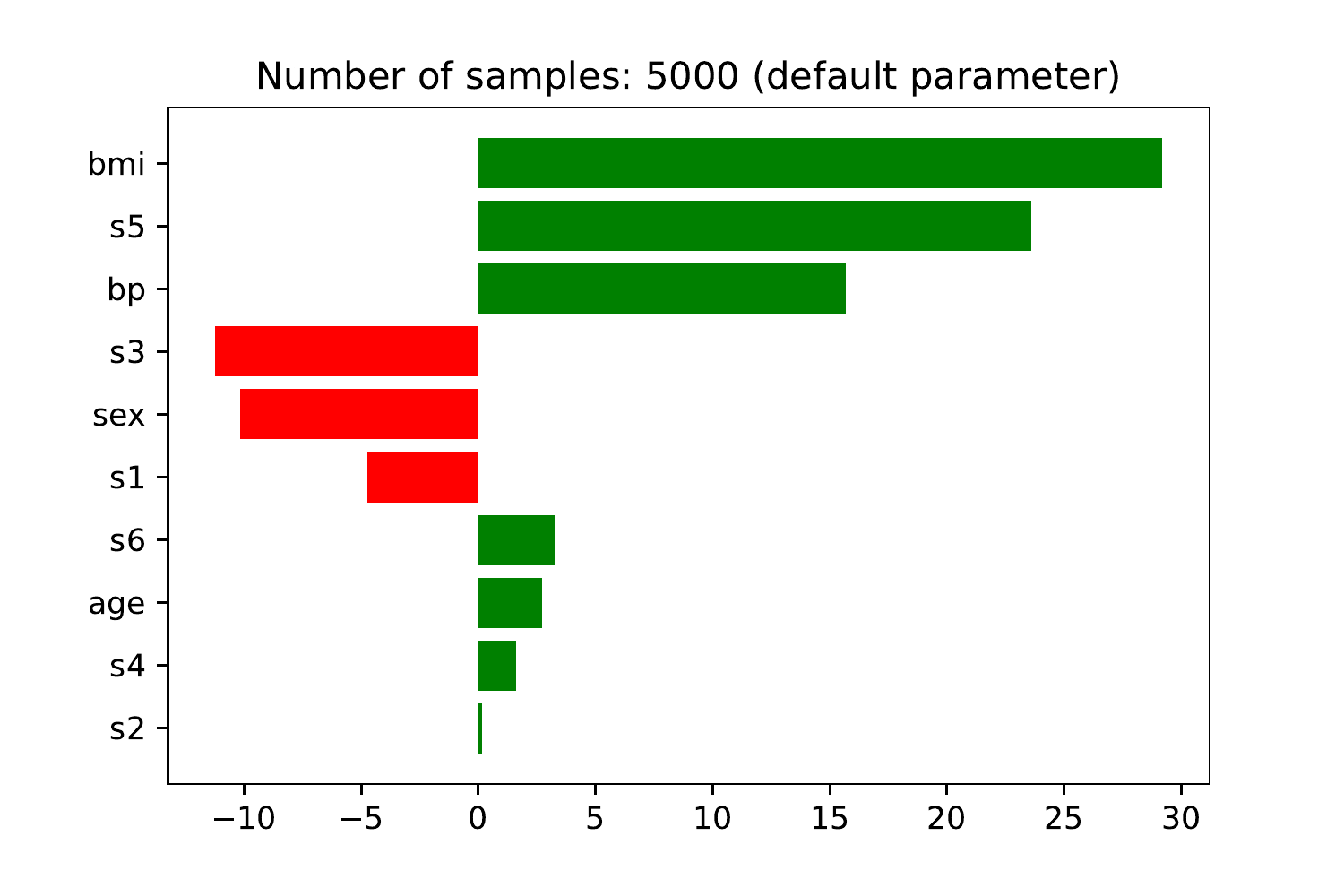}
\caption{Different sizes of explanations produced by LIME for an observation from Diabetes dataset}
\label{fig:explanations}
\end{figure}

Compactness has an impact on the other properties \cite{nauta2022anecdotal}. Moreover, Bob, the physician, should observe the explanations to confirm the number of features that are necessary to understand the prediction. Thus, it is appropriate to compare the XAI solutions using the aggregated score while taking into account the number of features. 
For that, let's pick a particular observation from the Diabetes dataset. The model makes a prediction and Bob asks \textit{Why this prediction?} and wants to know the features contributing to this prediction. The explanations produced by the XAI solutions recommended by AutoXAI are in Figure \ref{fig:explanations}. At the bottom right is LIME with default hyperparameters. We can see that some features have little influence on the prediction and are useless to answer the question of Bob. With these different-sized explanations, Bob can see what is important and what is negligible to him. He can thus choose the size of explanation he wants, keeping in mind the scores of the properties and the aggregated score.

\subsection{SPAM detection}

For the second use case, let's first describe the context. We consider a computer security manager in a laboratory, Charli. The staff complains about receiving spam in the chat service of the laboratory. Charli manages this service and knows little about ML, Charli wants to use an ML algorithm to automatically detect spam. Therefore, Charli finds an off-the-shelf GloVe embedding \cite{pennington2014glove} and implement a LSTM \cite{10.1162/neco.1997.9.8.1735}. Charli trains the model on a spam dataset, obtains a good accuracy, and decides to try it one week on the chat service. At the end of this week, a colleague of Charli shows a message that has not been sent because suspected as a spam. Therefore, Charli ends up wondering "Where does the model fail ?", more specifically, "What do false positives and false negatives look like?". To answer these questions, Charli wants to implement an XAI solution. In the end, Charli wants to use this knowledge to be able to explain to users the decisions made with their data and possibly reduce the error rate.

Charli needs examples of messages with specific predictions to know what they look like.
Therefore, we define $\mathcal{E}_{i'}=$\textit{What kind of data lead to this prediction?} and $\mathcal{E}^\prime_{j'}=$\textit{data points as explanations} (also called prototypes).

For the constraints of the context, Charli does not want to miss any type of message leading to an error. Ideally, Charli wishes that each data point should have a similar prototype (completeness). However, Charli also wants to avoid having too many prototypes and avoid redundancies (compactness). 
\begin{table*}[]
\caption{Extract from the ranking produced by AutoXAI on SMS Spam dataset}
\centering
\begin{tabular}{c|ccc|cc}
\hline
Aggregated & Scaled & Scaled & Scaled & XAI & Hyperparameters              
\\score & Representativeness & Diversity & NoP & solution &
\\ \hline
0.483            & 0.904                     & 0.248            & -0.303     & k-medoids    & heuristic;300;pam;cosine;16  \\
0.466            & 0.463                     & 0.412            & 0.030      & MMD-critic   & 1.000;13                     \\
0.384            & 0.224                     & 0.201            & 0.251      & Protodash    & gaussian;11.90;11            \\
0.367            & -0.660                    & 0.589            & 0.917      & MMD-critic   & 1.000;5                      \\
0.331            & -0.444                    & 0.048            & 0.917      & k-medoids    & build;224;alternate;cosine;5 \\
0.255            & -0.580                    & 0.092            & 0.917      & Protodash    & gaussian;21.43;5            
\end{tabular}
\label{table:results_spam}
\end{table*}

The dataset used is derived from the UCI SMS Spam dataset \cite{Dua:2019}, it has 8714 features for 5572 data samples and was built using TFIDF \cite{jones1972statistical}. Charli separates the dataset into 4 subsets according to the model results: true positives, true negatives, false positives, and false negatives. Thus, the ML model (GloVe and LSTM) is no longer necessary for this experiment.
The implemented XAI solutions are MMD-critic \cite{kim2016examples}, Protodash \cite{gurumoorthy2019efficient} and k-medoids \cite{kaufman1990partitioning}.

MMD-critic \cite{kim2016examples} proposes prototypes as explanations. To accurately represent the data distribution, it minimizes the discrepancy between the prototypes distribution and the data distribution. It also proposes critics which are points that are not well represented by the prototypes. Protodash \cite{gurumoorthy2019efficient} generalizes \cite{kim2016examples}, it is a fast prototype selection that also associates non-negative weights to prototypes which are indicative of their importance. Eventually, although it was not proposed for XAI, we use k-medoids \cite{kaufman1990partitioning} as it is a baseline that gives comparable results. It finds medoids (prototypes here) such that the distance between one prototype and the other points of its data group is minimal. 

The implemented XAI evaluation metrics and their corresponding properties are:
\begin{itemize}
    \item \textbf{Non-representativeness} \textit{Completeness}
    \item \textbf{Diversity} \cite{nguyen2020quantitative} \textit{Compactness} (\textit{redundancy})
    \item \textbf{Number of prototypes} \textit{Compactness} (\textit{size})
\end{itemize}

We propose a new \textbf{non-repre\-sen\-ta\-ti\-ve\-ness} metric to assess whether there is a close prototype for each data sample on average. Unlike \cite{nguyen2020quantitative}, it is model-agnostic, as XAI solutions that do not use an ML model should not be evaluated according to the ML model. To assess the \textit{Compactness}, we use the \textbf{number of prototypes} for the size of the explanation and \textbf{diversity}. For \textbf{diversity}, we adapt \cite{nguyen2020quantitative} proposal to measure the mean distance between prototypes. As \textbf{diversity} and \textbf{number of prototypes} are fundamentally different, we consider that they correspond to two different sub-properties (\textit{redundancy} and \textit{size} respectively) and let the user give a weight to each.
See Table \ref{tab:evaluation} for the formulas.

For the aggregation in this scenario, Charli sets the weights to 2, 1, 2 for \textbf{non-representativeness}, \textbf{diversity} and \textbf{number of prototypes} respectively. Charli, sets the number of epochs to 25. The HPO strategy is also the Gaussian Process here.

An extract from the ranking produced by AutoXAI for SMS Spam dataset is in Table \ref{table:results_spam}. As previously, the XAI solutions are sorted in descending order according to the Aggregated score. For each XAI solution, we present 2 results, the overall best score and the best score for a number of prototypes lower or equal to 5. In the Hyperparameter columns, k-medoids has the following hyperparameters: the initialization method, the maximum number of iterations, the algorithm to use, the metric and the number of medoids to generate \footnote{\url{https://scikit-learn-extra.readthedocs.io/en/stable/generated/sklearn_extra.cluster.KMedoids.html}}. MMD-critic has the following hyperparameters: the gamma value and the number of prototypes to find. Protodash has the following hyperparameters: the kernel to use, the value of sigma, and the number of prototypes to find.

\begin{table*}[]
\caption{Computation time and scores for LIME evaluation without and with the time-saving evaluation strategies}
\resizebox{\textwidth}{!}{
\begin{tabular}{ccccccccc}
\cmidrule[\heavyrulewidth]{2-9}
           & \multicolumn{2}{c}{No strategy}    & \multicolumn{2}{c}{Early stopping} & \multicolumn{2}{c}{Information sharing} & \multicolumn{2}{c}{Both strategies} \\ \cline{2-9} 
           & Time   & Score                     & Time            & Score            & Time               & Score              & Time            & Score             \\ \hline
Robustness & 488.04 $\pm$ 14.79 & -70.67 $\pm$ 0.64 & 17.45 $\pm$ 1.37          & -67.41 $\pm$ 3.26         & 34.15 $\pm$ 1.70            & -69.22 $\pm$ 2.24 & 1.22 $\pm$ 0.13 & -67.62 $\pm$ 3.12\\
Infidelity & 11.84$\pm$ 0.27 & -5.29 $\pm$ 0.06 & 0.46 $\pm$ 0.05 & -5.00 $\pm$ 0.78 & 9.32 $\pm$ 0.37 & -5.40 $\pm$ 0.12 & 0.35 $\pm$ 0.03 & -5.48 $\pm$ 0.6\\ \bottomrule
\multicolumn{9}{l}{\small{Computation times are in seconds. For evaluation scores, the higher, the better.}}
\end{tabular}}
\label{table:time-saved}
\end{table*}

\begin{figure}
\begin{subfigure}{.5\textwidth}
\centering
\includegraphics [width=\textwidth]{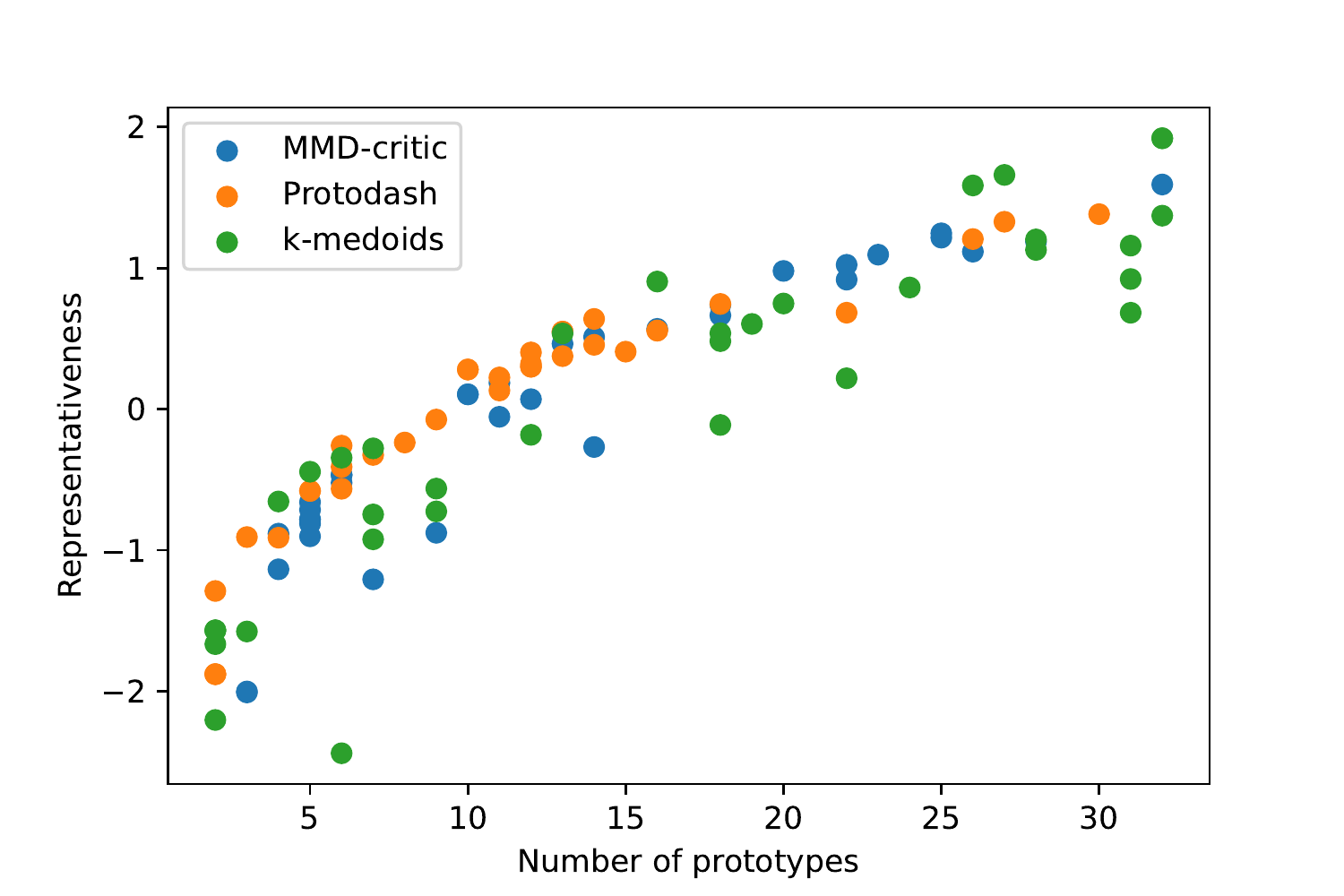}
\caption{Influence of the number of prototypes on representativeness}
\label{fig:rpz_nop}
\end{subfigure}
~
\begin{subfigure}{.5\textwidth}
\centering
\includegraphics [width=\textwidth]{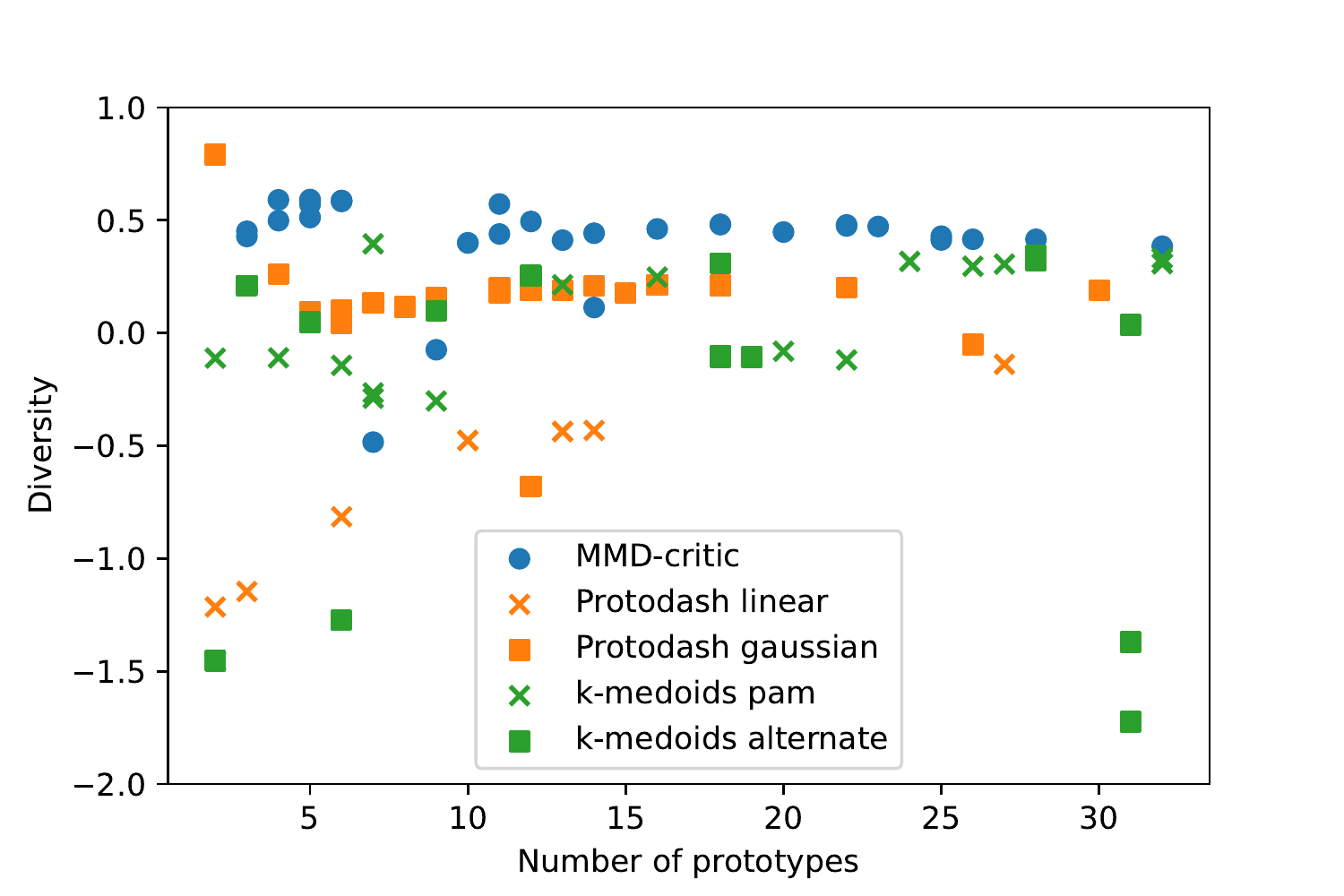}
\caption{Influence of the number of prototypes on diversity}
\label{fig:div_nop2}
\label{div_nop_main}
\end{subfigure}
\caption{Influence of properties on each others for SMS Spam dataset}
\end{figure}

% \begin{figure}
%     \begin{subfigure}{.5\textwidth}
%         \includegraphics [width=\textwidth]{figures/diabetes_scores.pdf}
%         \caption{Results for Diabetes dataset}
%         \label{fig:diab_scores}
%     \end{subfigure}
%   ~ 
%     \begin{subfigure}{.5\textwidth}
%         \includegraphics [width=\textwidth]{figures/pima_scores.pdf}
%         \caption{Results for Pima indians dataset}
%         \label{fig:pima_scores}
%     \end{subfigure}
%     \caption{Robustness loss and Fidelity loss}
% \end{figure}

Regarding Table \ref{table:results_spam}, we observe that k-medoids has the best Aggregated score with high representativeness. We also observe that the representativeness score is systematically lower with fewer prototypes. This trend is shown in Figure \ref{fig:rpz_nop}. It can be seen that representativeness is more important with more prototypes. Indeed, intuitively, the more prototypes there are, the more likely it is that a data point is close to one of them. It results in a trade-off between compactness and completeness that encourage choosing appropriate weights for the properties.
In Table \ref{table:results_spam}, we observe that the two algorithms for k-medoids (pam and alternate) are performing well, while Protodash seems to have better results with the Gaussian kernel. This is confirmed in Figure \ref{fig:div_nop2} where Protodash performs better in terms of diversity with Gaussian kernel while k-medoids have equivalent results with pam and alternate algorithm. MMD-critic regularly has the best scores in diversity and k-medoids has a higher variance in scores on both diversity and representativeness.

Using the best scoring XAI solution on the SMS Spam dataset, Charli obtains the following prototypes for false positives:
\begin{itemize}
    \item Hey pple...\$700 or \$900 for 5 nights...Excellent location wif breakfast hamper!!!
    \item Unlimited texts. Limited minutes.
\end{itemize}
As well as the following protoypes for false negatives:
\begin{itemize}
    \item FROM 88066 LOST £12 HELP
    \item Money i have won wining number 946 wot do i do next
\end{itemize}
With these representative examples, Charli can explain to users what kind of messages the model may misclassify and can work on the data and the ML model while tracking predictions for these messages.

\subsection{Time-saving evaluation strategies}
Preliminary results for time-saving evaluation strategies (see Section \ref{section:time}) are obtained on the first use case (see Section \ref{subsection:expe_diab}) with diabetes dataset and MLPRegressor model. The XAI solution is LIME with default hyperparameters and the XAI evaluation metrics are robustness and infidelity. AutoXAI is run on a laptop with a 2.40 GHz octa-core CPU. Table \ref{table:time-saved} shows the average with its standard deviation of computation time and evaluation score. Early stopping saves 96.42\% of the time for robustness and 96.13\% for infidelity.
Information sharing saves 93\% of the time for robustness and 21.26\% of the time for infidelity. For Information sharing with robustness, the maxima used for computing scores are obtained with LIME with other hyperparameters, hence the scores difference. With infidelity, the generated perturbation points and their corresponding predictions are obtained with another seed which explains the small difference in score compared to the no strategy baseline.
Using both strategies saves 99.75\% of the time for robustness and 97\% for infidelity.

\section{Conclusion and perspectives}\label{section:Conclusion}
In this paper, we propose AutoXAI, a framework that recommends the best XAI solutions according to the context of its user. AutoXAI automates the tedious task of selecting an XAI solution and its hyperparameters. It produces a ranking of solutions taking into account the preferences of the user. 
Although AutoXAI is system-centric like AutoML, here the user specifies the needs, the constraints and chooses the XAI solution in the ranking.
It saves time and does not require the user to have deep knowledge of XAI. AutoXAI can also serve a researcher who wants to test an XAI proposal and monitor results. Indeed, as AutoXAI tests multiple XAI evaluation metrics on multiple XAI solutions and hyperparameters, potential relationships could be discovered. In this paper, we show there might be a trade-off between properties. Compactness, in particular, should be monitored and the user should decide by checking the explanations.
The choice of one explanation over another one may raise ethical issues. Indeed, encouraging too much particular properties of explanations can lead to bias. For instance, reducing the number of prototypes can lead to missing the less represented data samples. Alternatively, a small number of features in an explanation can hide a bias in a model. 

A short-term work could complement AutoXAI with new adaptations of AutoML methods. Longer perspectives should be to apply AutoXAI in a real-world setting and analyze users' feedback to assess its usefulness and give opportunities to improve it. Lastly, studying the influence of XAI properties on each other will be an important topic of study in the field of XAI solution evaluation.

\bibliographystyle{unsrtnat}
\bibliography{autoxai}  %%% Uncomment this line and comment out the ``thebibliography'' section below to use the external .bib file (using bibtex) .





% \end{thebibliography}

\end{document}